\documentclass[11pt,a4paper]{article}
\usepackage[hyperref]{emnlp-ijcnlp-2019}
\usepackage{times}
\usepackage{latexsym}

\usepackage{graphicx}
\usepackage{url}

\aclfinalcopy 

\newcommand\conforg{SIGDAT}

\title{GeoSQA: A Benchmark for Scenario-based Question Answering in the Geography Domain at High School Level}

\author{Zixian Huang, Yulin Shen, Xiao Li, Yuang Wei, Gong Cheng, Lin Zhou, Xinyu Dai, Yuzhong Qu \\
  National Key Laboratory for Novel Software Technology, Nanjing University, China \\
  {\tt \{zixianhuang,ylshen,xiaoli,weiyuang\}@smail.nju.edu.cn,} \\
  {\tt \{gcheng,daixinyu,yzqu\}@nju.edu.cn,zhoul@nlp.nju.edu.cn} \\}

\date{}

\begin{document}
\maketitle
\begin{abstract}
Scenario-based question answering (SQA) has attracted increasing research attention. It typically requires retrieving and integrating knowledge from multiple sources, and applying general knowledge to a specific case described by a scenario. SQA widely exists in the medical, geography, and legal domains---both in practice and in the exams. In this paper, we introduce the GeoSQA dataset. It consists of 1,981~scenarios and 4,110~multiple-choice questions in the geography domain at high school level, where diagrams (e.g.,~maps, charts) have been manually annotated with natural language descriptions to benefit NLP research. Benchmark results on a variety of state-of-the-art methods for question answering, textual entailment, and reading comprehension demonstrate the unique challenges presented by SQA for future research.
\end{abstract}

\section{Introduction}

Scenario-based question answering (SQA) is an emerging application of NLP~\cite{watsonpaths}. Different from traditional QA, a question in SQA is accompanied by a \emph{scenario}, e.g., a patient summary in the medical domain asking for diagnosis or treatment. A scenario differs from a document given in the reading comprehension task where the answer can be extracted or abstracted from the document~\cite{squad,msmarco,race}. SQA requires retrieving and integrating knowledge from multiple sources, and applying general knowledge to a specific case described by the scenario.

\begin{figure*}
\centering
\includegraphics[width=\textwidth]{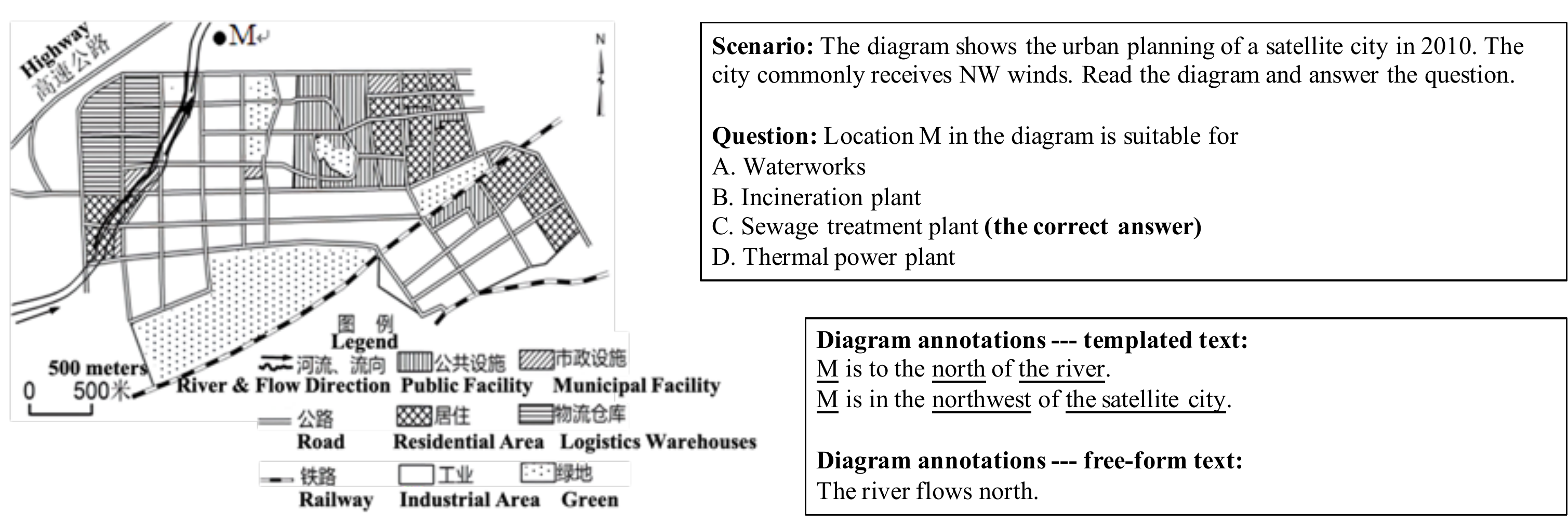}
\caption{An example of a scenario, a question, and diagram annotations.}
\label{fig:example}
\end{figure*}

SQA has found application in many fields, especially in the legal domain~\cite{legalnaacl,legalemnlp17,legalemnlp18} and in high-school geography exams~\cite{gaokaoeswc,gaokaoccks}. The latter is particularly challenging because a geographical scenario consists of both text and diagrams (e.g.,~maps, charts). Questions include city planning, climates, agriculture planning, transportation, etc. An example of a scenario and a question is presented in Figure~\ref{fig:example}.

Geographical SQA has posed great challenges to NLP and related research, ranging from scenario understanding to cross-modal knowledge integration and reasoning. However, there is a lack of large datasets and benchmarking efforts for this task. In this paper, we introduce GeoSQA---an SQA dataset in the geography domain consisting of 1,981~scenarios and 4,110~multiple-choice questions at high school level. In particular, each diagram has been manually annotated with a high-quality natural language description of its content, as illustrated in Figure~\ref{fig:example}. This labor-intensive effort significantly extends the use of GeoSQA, which can support visual SQA (using the diagrams), natural language based SQA (using the annotations of diagrams), and even the diagram-to-text research. We test the effectiveness of a variety of methods for question answering, textual entailment, and reading comprehension on GeoSQA. The results demonstrate its unique challenges, waiting for more effective solutions.

The remainder of the paper is organized as follows. Section~\ref{sect:rw} discusses related work. Section~\ref{sect:ds} describes the GeoSQA dataset. Section~\ref{sect:exp} reports benchmark results. Section~\ref{sect:concl} concludes the paper.

\section{Related Work}\label{sect:rw}

\subsection{Scenario-based Question Answering}
Scenario-based question answering (SQA) is introduced by Lally et al.~\shortcite{watsonpaths}, where the WatsonPaths system is presented to answer questions that describe a medical scenario about a patient and ask for diagnosis or treatment.
SQA also finds application in the legal domain, where a legal case describes a scenario to be decided~\cite{legalnaacl,legalemnlp17,legalemnlp18}.

For some domains, reasoning with domain knowledge is essential to SQA. Therefore, such questions often appear in exams like China's version of the SAT called Gaokao. For example, for the geography domain, Ding et al.~\shortcite{gaokaoeswc} and Zhang et al.~\shortcite{gaokaoccks} construct a knowledge graph to support answering scenario-based geography questions at high school level.

\subsection{Related Datasets}
There are many datasets for traditional QA, such as WebQuestions~\cite{webquestions} and WikiQA~\cite{wikiqa}. A closely related task is reading comprehension, where the answer to a question is extracted or abstracted from a given document~\cite{squad,msmarco,race}.
By comparison, SQA is arguably more difficult because a scenario is present and contextualizes a question, but no direct answer can be identified from the scenario.

The GeoSQA dataset introduced in this paper is not the first resource for geographical SQA. Ding et al.~\shortcite{gaokaoeswc} and Zhang et al.~\shortcite{gaokaoccks} have made their datasets public. However, compared with GeoSQA, their datasets are small and, more importantly, they ignore diagrams which represent a unique challenge to geographical SQA. By contrast, diagrams are included in GeoSQA for completeness, and have been manually annotated with natural language descriptions for extended use---including but not limited to NLP research.

Existing SQA datasets for other domains include the TREC Precision Medicine track~\cite{trec} for the medical domain, and CAIL~\cite{cail} for the legal domain. However, SQA in the geography domain requires different forms of knowledge and different reasoning capabilities, and has posed different research challenges.
\section{The GeoSQA Dataset}\label{sect:ds}
GeoSQA is an SQA dataset in the geography domain, containing 1,981~scenarios and 4,110~multiple-choice questions at high school level. A scenario consists of a piece of text and a diagram, supporting 1--5~questions. A diagram is annotated with a natural language description of its content. A question has four options that are possible answers. Exactly one of them is the correct answer. The dataset is available online\footnote{\url{ws.nju.edu.cn/gaokao/geosqa/1.0/}}.

\subsection{Data Collection and Deduplication}
We crawled over 6,000~scenarios and 13,000~questions from Gaokao and mock tests that are available on the Web.
However, some scenarios are just copies or trivial variants of others. There is a need to clean and deduplicate the collected data.

\textbf{Method.}
The problem is to decide whether a pair of scenarios are (near) duplicates or not.


We firstly establish a matching between their structures. The matching consists of 6~pairs of their text elements: 1~pair of their scenario text, 1~pair of their most similar questions, and 4~pairs of the most similar options of the above two questions. Text similarity is computed by the cosine similarity between two bags of words.

Then we extend a popular text matching method called MatchPyramid~\cite{MatchPyramid} to classify a pair of scenarios as duplicates or not. The original implementation of MatchPyramid can only process a pair of text. We extend it to process all the 6~pairs of text by concatenating their feature vectors inside MatchPyramid.

\textbf{Experiments.}
To evaluate our method, we manually label 1,000~pairs of scenarios where positive and negative examples are balanced. The set is divided into training, validation, and test sets with a 60-20-20 split. Our method achieves an accuracy of~95.3\% on the test set, showing its satisfying performance.

Then we apply our method to the entire dataset. We index all the scenarios using Apache Lucene. For each scenario, we retrieve 10~top-ranked scenarios as suspect duplicates. Each pair of scenarios is classified by our method, which is trained using all the 1,000~labeled examples.

To verify the quality of the final results, we randomly sample and manually check 100~pairs of scenarios that are predicted to be duplicates. Indeed, all of them are decided correctly. We also randomly sample 50~scenarios and, for each of them, we retrieve and manually check 10~top-ranked scenarios that are predicted to be non-duplicates. Only~6\% are decided incorrectly, suggesting a low degree of redundancy of our data.
 
\subsection{Diagram Annotation}
Crawled diagrams are images. To extend the use of GeoSQA and to better support NLP research, we manually annotate each diagram with a high-quality natural language description of its content so that NLP researchers can use these text annotations instead of the original diagrams.

\textbf{Annotation.}
We recruited 30~undergraduate students from one of the top-ranked universities in China as annotators. All of them had an excellent record in geography during high school.

Each diagram is assigned to one annotator, who also has access to the scenario text and related questions. The annotator firstly categorizes the diagram according to a hierarchy of categories. Each category is associated with a set of text templates that are recommended to be used in annotations as far as possible.
However, the annotator is free to use any form of text to annotate information that is not covered by the provided templates.

Annotations are required to precisely reflect the content of the diagram. All the information related to every supported question and every option should be annotated. On the other hand, inferring new knowledge via human reasoning is prohibited.

Note that the entire annotation process is designed to be iterative. The 22~diagram categories and 81~text templates are not predefined but incrementally induced during the experiment. However, there are still 11\%~of the diagrams that are believed to not belong to any category. No templates are provided for their annotations.

An example of annotations is shown in Figure~\ref{fig:example}.

\textbf{Audit.}
To ensure the quality of the annotations, we recruited 3~senior annotators to audit the results. Each diagram is audited by one senior annotator, who rates the annotations from three dimensions in the range of 1--5.
\begin{itemize}
    \item Sufficiency: The annotations cover all the necessary information in the diagram that is useful for answering related questions.
    \item Fairness: The annotations are not biased towards any particular option of a question.
    \item Objectiveness: The annotations are plain descriptions of the diagram---not influenced by human reasoning.
\end{itemize}
\noindent The scenarios where the annotations of the diagram are rated below~3 in any dimension are excluded from the dataset.


\section{Benchmark Results}\label{sect:exp}
\label{sect:1}
We tested several state-of-the-art methods for question answering, textual entailment, and reading comprehension on our GeoSQA dataset.

\subsection{Corpora}
We use two corpora as background knowledge. \textbf{Textbooks} contains 15K~sentences extracted from two high-school geography textbooks. \textbf{Wikipedia} contains 1M~articles in the latest Chinese edition of Wikipedia. We index their sentences using Apache Lucene.

\subsection{Methods}
We tested two text matching methods. In \textbf{IR}~\cite{SearchPMI}, for each option, we use a combination of the scenario text, the question, and the option as a query, to retrieve the top-ranked sentence from a corpus. We use the ranking score of this sentence as the score of the option. Finally, we choose the option with the highest score as the answer. In \textbf{PMI}~\cite{SearchPMI}, for each option, we calculate the Pointwise Mutual Information (PMI) between the question and the option as the score of the option. Finally, we choose the option with the highest score as the answer. Probabilities in PMI are estimated based on a corpus. 

We tested four textual entailment methods: \textbf{ESIM}~\cite{ESIM}, \textbf{DIIN}~\cite{DIIN}, \textbf{BERT$_{NLI}$}~\cite{bert}, and \textbf{BiMPM}~\cite{BiMPM}.
The first three methods were trained on the XNLI dataset~\cite{XNLI}.
The last method was trained on the LCQMC dataset~\cite{LCQMC}.
For each option, a textual entailment method retrieves six top-ranked sentences from a corpus to form the entailing text. Retrieval follows the procedure described in the above-mentioned \textbf{IR} method. The scenario text and diagram annotations may or may not be included in the entailing text, depending on the configuration. A combination of the question and the option form the entailed text. Finally, we choose the option with the highest entailment score as the answer.

We tested one reading comprehension method: \textbf{BERT$_{RC}$}~\cite{bert}. It was trained on the DuReader dataset~\cite{DuReader}.
For each option, a reading comprehension method retrieves six top-ranked sentences from a corpus as part of the passage for reading comprehension. Retrieval follows the procedure described in the above-mentioned \textbf{IR} method. The scenario text and diagram annotations may or may not be included in the passage, depending on the configuration. Finally, the reading comprehension method extracts a text span from the passage. We choose the option that is the most similar to the extracted text span as the answer. Similarity is computed by the cosine similarity between two bags of words.






\subsection{Results}
The results are summarized in Table~\ref{table:baseline}. Note that a question has four options. Even guessing randomly, the expected proportion of correctly answered questions would be 25\%.

Almost all the methods performed similar to random guess, showing that SQA on our dataset has its unique challenges.

\begin{table}[t!]
\begin{center}
\begin{tabular}{lcc}
\hline & Textbooks & Wikipedia \\ \hline
IR & 25.24 & 25.14 \\
PMI & 26.22 & 25.19 \\
\hline
ESIM w/o scenario & 25.85 & 25.41 \\
ESIM w/ scenario  & 24.34 & 24.41 \\
DIIN w/o scenario & 24.15 & 25.20 \\
DIIN w/ scenario & 25.11 & 24.89 \\
BERT$_{NLI}$ w/o scenario & 24.29 & 24.17 \\
BERT$_{NLI}$ w/ scenario & 24.97 & 24.68 \\
BiMPM w/o scenario & 24.13 & 24.51 \\
BiMPM w/ scenario & 24.76 & 23.81 \\
\hline
BERT$_{RC}$ w/o scenario & 24.81 & 24.78\\
BERT$_{RC}$ w/ scenario & 23.66 & 23.01 \\
\hline
\end{tabular}
\end{center}
\caption{\label{table:baseline} Proportions of correctly answered questions.}
\end{table}

\subsection{Discussion}
To explain the poor performance of existing methods, we have identified the following challenges. First, SQA relies on domain knowledge that is not provided in the scenario. However, relevant knowledge may fail to be retrieved from the corpus. Second, for some questions, commonsense knowledge is needed but is not included in textbooks and may fail to be retrieved from Wikipedia. Third, the retrieved general knowledge needs to be applied to the specific case described by a scenario. Existing QA and reading comprehension methods hardly have this capability.
\section{Conclusion}\label{sect:concl}
We have contributed GeoSQA---a large SQA dataset where diagrams are present and have been manually annotated with natural language descriptions. We have tested a variety of existing methods on our dataset. The results are not satisfactory, thus demonstrating the unique challenges presented by the SQA task on our dataset. In future work, we will work towards more effective solutions to meet the challenges.

Researchers are invited to use GeoSQA to support their own tasks, including but not limited to natural language based SQA, visual SQA, and the diagram-to-text task.

\section*{Acknowledgments}
This work was supported in part by the National Key R\&D Program of China under Grant 2018YFB1005100, and in part by the NSFC under Grant 61572247. Gong Cheng was supported by the Qing Lan Program of Jiangsu Province. We would like to thank all the annotators.

\bibliography{emnlp-ijcnlp-2019}
\bibliographystyle{acl_natbib}

\end{document}